\documentclass{article}
\usepackage{spconf,amsmath}
\usepackage[pdftex]{graphicx}
\usepackage{bm}
\usepackage{url}
\usepackage{amssymb}
\usepackage{cite}
\usepackage{color}

\newcommand{\argmin}{\mathop{\rm arg~min}\limits}

\title{AdaFlow: Domain-Adaptive Density Estimator with Application to Anomaly Detection and Unpaired Cross-Domain Translation}

\name{
Masataka Yamaguchi$^{1}$,
Yuma Koizumi$^{2}$, and
Noboru Harada$^{2}$
}
\address{
$^{1}$: \normalsize{NTT Communication Science Laboratories, Kanagawa, Japan}\\
$^{2}$: \normalsize{NTT Media Intelligence Laboratories, Tokyo, Japan}
}

\begin{document}
\ninept
\maketitle

\begin{abstract}
We tackle unsupervised anomaly detection (UAD), a problem of detecting data that significantly differ from normal data.
UAD is typically solved by using density estimation. Recently, deep neural network (DNN)-based density estimators, such as Normalizing Flows, have been attracting attention.
However, one of their drawbacks is the difficulty in adapting them to the change in the normal data's distribution.
To address this difficulty, we propose {\it AdaFlow}, a new DNN-based density estimator that can be easily adapted to the change of the distribution.
AdaFlow is a unified model of a Normalizing Flow and Adaptive Batch-Normalizations, a module that enables DNNs to adapt to new distributions.
AdaFlow can be adapted to a new distribution by just conducting forward propagation once per sample; hence, it can be used on devices that have limited computational resources.
We have confirmed the effectiveness of the proposed model through an anomaly detection in a sound task.
We also propose a method of applying AdaFlow to the unpaired cross-domain translation problem, in which one has to train a cross-domain translation model with only unpaired samples. We have confirmed that our model can be used for the cross-domain translation problem through experiments on image datasets.
\end{abstract}


\vspace{-3pt}

%
%
%
%
%

\section{Introduction}
\label{sec:intro}

Anomaly detection, also known as outlier detection, is a problem of detecting data that significantly differ from normal data~\cite{Hodge_2004,Patcha_2007,ASD_survey}.
Since such anomalies might indicate symptoms of mistakes or malicious activities, their prompt detection may prevent such problems.
Therefore, anomaly detection has received much attention and been applied for various purposes.

In this paper, we specifically consider unsupervised anomaly detection (UAD), in which only normal data can be used for training anomaly detection models.
UAD is typically solved by first training a normal model with normal data and then estimating the deviance of each testing sample with the trained model.
In the anomaly detection field, many types of normal models have been investigated. In the early studies, a Gaussian distribution was used~\cite{shewhart1931economic,abraham1979bayesian}, and recently, more flexible statistical models have been used such as a Gaussian mixture model (GMM)~\cite{agarwal2007detecting, Koizumi_2017_ADS}.
More recently, deep neural network (DNN)-based methods have been investigated such as an Auto-Encoder (AE)~\cite{zhou2017anomaly, Koizumi_2018_IEEE_ADS}, a Variational Auto-Encoder (VAE)~\cite{kingma2013auto,VAE_Anomaly,Kawachi_2018}, and Generative Adversarial Networks (GAN)~\cite{goodfellow2014generative,schlegl2017unsupervised, lim2018doping}.

In the typical setting of UAD, one assumes that training and testing data are sampled from the same distribution.
However, this assumption does not hold in certain practical scenarios.
Let us consider the anomaly detection problem on facility equipments.
Typically, such equipments have various operation patterns, and the environmental noise patterns around them may change due to certain factors such as seasons and the weather.
In this case, the above assumption does not always hold; hence, simply applying existing normal models to such problems may significantly decrease the anomaly detection accuracy. 
A na\"ive method one can use to avoid this is to adapt normal models to a new distribution by conducting fine-tuning with newly-collected normal data.
However, fine-tuning requires high memory and computational costs and cannot be easily conducted with devices installed in facility equipments that typically have only limited computational resources.
Therefore, a more efficient adaptation method is needed.

To address this problem, we propose a new density estimator named {\it AdaFlow}, a unified model of Normalizing Flows (NFs)~\cite{rezende2015variational,dinh2016density}, a powerful DNN-based density estimator, and the Adaptive Batch Normalization (AdaBN)~\cite{li2016revisiting}, a module that enables DNNs to handle different domains' data. 
AdaBN alleviates the difference between domains by scaling and shifting each domain's input data so that each domain's mean and variance are zero and one, respectively.
Since AdaBN can be adapted to a new domain by just adjusting its statistics with the domain's data, the adaptation step of AdaFlow can be done by just conducting forward-propagation only once per sample.
Therefore, AdaFlow can be used on devices that have limited computational resources.

We also propose a method of applying AdaFlow to the unpaired cross-domain translation problem, in which one has to train a cross-domain translation model with only unpaired data. We show the effectiveness of using AdaFlow for this problem through cross-domain translation experiments on image datasets.

\vspace{-4pt}

\section{Related Work}
\label{sec:conv}

\subsection{Unsupervised anomaly detection}
\label{sec:unsupervised}

In UAD, the deviation between a normal model and observation is computed; the deviation is often called the ``{\it anomaly score}''.
One way of computing anomaly scores is a density estimation-based approach. This approach first trains a density estimator $q_{\theta} (\cdot) $, such as a Gaussian distribution function, with normal data, and then computes the negative log-likelihood of each testing data $\bm{x} \in \mathbb{R}^{D}$ with $q_{\theta} (\cdot) $. In this approach, its negative log-likelihood is used as its anomaly score $\mathcal{A}(  \bm{x}, \theta )$, i.e.,
\begin{equation}
\mathcal{A}(  \bm{x}, \theta )   =
-\ln q_{\theta} (\bm{x} ).
\end{equation}
Then, $\bm{x}$ is determined to be anomalous when the anomaly score exceeds a pre-defined threshold $\phi$:
\begin{equation}
\mathcal{H}(\bm{x}, \phi) =
 \begin{cases}
 0 \mbox{ }(\mbox{Normal}) & \mathcal{A} (\bm{x}, \theta ) < \phi \\
 1 \mbox{ }(\mbox{Anomaly})& \mathcal{A} (\bm{x}, \theta ) \ge \phi
 \end{cases}.
\label{eq:hard_thres}
\end{equation}

Recently, deep learning has also been investigated for defining normal models for UAD.
Several studies on deep-learning-based UAD employed an AE~\cite{zhou2017anomaly, Koizumi_2018_IEEE_ADS} (or a VAE \cite{VAE_Anomaly,Kawachi_2018}).
The AE-based anomaly detection framework defines the anomaly score as follows:
\begin{align}
\mathcal{A}(  \bm{x} , \theta )  &=
\lVert
\bm{x} - \mathcal{D}_{\theta_D}( \mathcal{E}_{\theta_E} (\bm{x} ) )
\rVert^2 ,
\label{eq:ae_anomalyscore}
\end{align}
where
$\lVert \cdot \rVert$ denotes the $L_2$ norm,
$\mathcal{E}$ and $\mathcal{D}$ are the encoder and decoder of the AE,
and $\theta_E$ and $\theta_D$ are its parameters, namely $\theta = \{ \theta_E, \theta_D \}$.
Then, $\theta$ is trained to minimize the anomaly scores of normal data as follows:
\begin{align}
\theta
&\gets \argmin_{\theta}
\frac{1}{N}
\sum_{n=1} ^{N}
\mathcal{A}(\bm{x}_n, \theta)
\label{eq:mmse_ae_sum},
\end{align}
where $\bm{x}_n$ is the $n$-th training sample and $N$ is the number of training samples.

Although it has been empirically shown that anomaly detection can be addressed by AE-based anomaly detection, one of its drawbacks is that there is no guarantee that minimizing Eq. (\ref{eq:mmse_ae_sum}) encourages anomaly scores of normal data to be less than those of anomaly data, because anomaly scores of anomaly data are not considered in Eq. (\ref{eq:mmse_ae_sum}).
In constrast, in the density estimation-based approach, minimizing NLLs of normal data encourages to maximize NLLs of the other data, including anomal data, since the integral value of the likelihood in the input space is always 1.
Therefore, instead of using the AE-based anomaly detection approach, we adopt the density estimation-based approach.
Specifically, in this paper, we adopt a Normalizing Flow (NF), a DNN-based flexible density estimator.
We explain its details in Section 3.

\subsection{Domain adaptation on DNN-based density estimator}

Although a DNN is a powerful tool for anomaly score computation, it may be problematic for practical use.
One problem occurs when adjusting the normal model to a new domain.
The distribution of normal data often varies due to aging of the target and/or change in environmental noise.
Therefore, we need to adapt the normal model to such fluctuations.
Let us formulate this problem.
Suppose that we have a normal model $q_{\theta}$ trained on $K \ge 1$ dataset(s) collected in individual domains.
When the distribution changes, we need to adapt $q_{\theta}$ to the new domain ($(K + 1)$-th domain) to obtain a new normal model $q'_{\theta}$.
This problem can be regarded as an analogy of {\it domain adaptation}~\cite{pan2010survey}.
Although several domain adaptation methods have been investigated~\cite{ganin2015unsupervised, bousmalis2016domain, saito2017asymmetric}, most require iterative optimization and huge memory, and such methods cannot be easily used with devices installed in most practical conditions, which typically have limited computational resources.
Therefore, in terms of the computational cost and required memory, a more efficient adaptation method is needed.

\vspace{-2.0mm}
\section{Proposed method}
\label{sec:prop}


\subsection{Normalizing Flow}
\label{sec:NormalizingFlow}

We adopt a Normalizing Flow (NF) as a density estimator.
NF represents a probabilistic density by transforming a base probabilistic density function $q_0(\bm{z}^{(0)})$ with a series of $M$ invertible projections $\{ f_{m} \}^M_{m=1}$ with each parameter $\{ \theta_m \}_{m=1}^M$.
In NF, $\bm{x}$ is regarded as a transformed variable with $\{ f_{m} \}^M_{m=1}$ as follows:
\vspace{-1.0mm}
\begin{equation}
\bm{x} = \bm{z}_M =  f_{M, \theta_{M}} \circ \cdots \circ f_{1, \theta_{1}} (\bm{z}^{(0)} );
\label{eq:nflow}
\end{equation}
thus, $\bm{z}^{(0)}$ can be obtained by the inverse transform of (\ref{eq:nflow}).
Following prior works~\cite{rezende2015variational,dinh2016density}, we employ a Gaussian distribution $\mathcal{N}(\bm{z}^{(0)} ; \bm{0}, \bm{I})$ for $q_0(\bm{z}^{(0)})$.
Then, the likelihood of the given sample $\bm{x}$ is obtained by repeatedly applying the rule for {\it change of variables} as follows:
\begin{align}
q_{\theta}( \bm{x} )  &= q_{0} ( \bm{z}^{(0)} ) \prod_{m=1}^{M} \left|  \frac{\partial f_{m, \theta_{m}}}{ \partial \bm{z}^{(m-1)} } \right|^{-1},
\end{align}
Thus, the anomaly score computed by NF can be expressed as
\begin{align}
\mathcal{A}(  \bm{x}, \theta )   &= -\ln q_{0} ( \bm{z}^{(0)} ) - \sum_{m=1}^{M} \ln \left|  \frac{\partial f_{m, \theta_{m}}}{ \partial \bm{z}^{(m-1)} } \right|^{-1}.
\end{align}
Parameters $\theta = \{ \theta_m \}_{m=1}^M$ can be trained by minimizing the anomaly scores as follows:
\vspace{-1.0mm}
\begin{equation}
\theta \gets \argmin_{\theta}
\sum_{k=1}^{K} \frac{1}{N_k} \sum_{n=1}^{N_k} \mathcal{A}(  \bm{x}_{n,k}, \theta ),
\label{eq:pre_train}
\end{equation}
where $\bm{x}_{n,k}$ and $N_k$ are the $n$-th training sample and the number of training samples of the $k$-th dataset, respectively.

\subsection{AdaFlow}
\label{sec:daflow}

\begin{figure}[ttt]
  \centering
\includegraphics[width=80mm,clip]{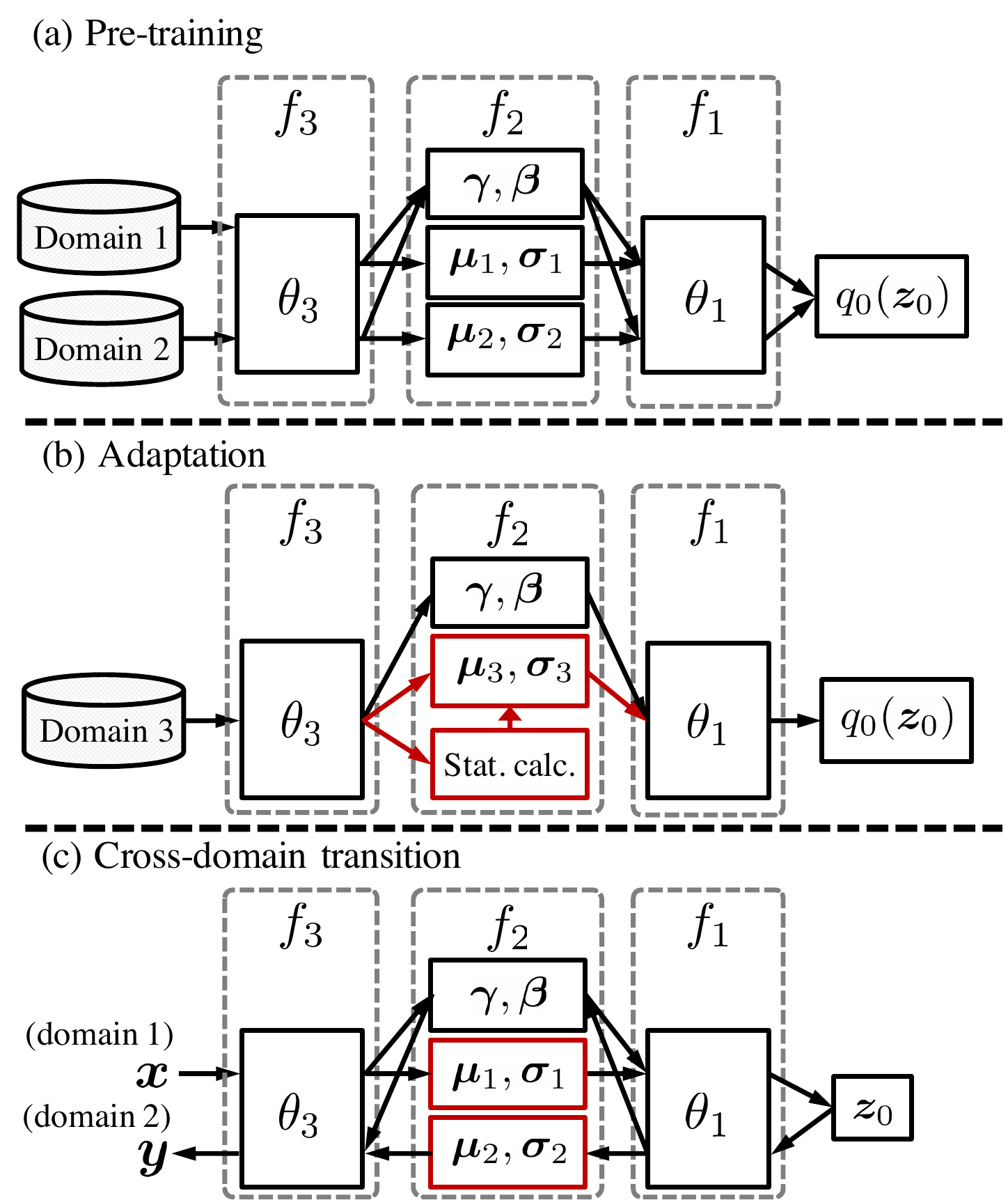}
  \caption{Simplified concept of AdaFlow; (1) pre-training, (2) adaptation, and (3) cross-domain transition.
In pre-training, all parameters $\{ \theta_m \}_{m=1}^3$ are trained with $K=2$ domain datasets.
For adaptation, BN statistics of second projection $f_2$ are computed from third domain dataset.
For cross-domain transition, BN statistics of input domain is used for inverse projection, and that of target domain is used for forward projection.
  }
  \label{fig:trn_flow}
  \vspace{-2.0mm}
\end{figure}

We consider domain adaptation for NF.
A na\"ive method of adapting NF to the $(K + 1)$-th dataset is to fine-tune all $\{ \theta_m \}_{m=1}^M$ with that dataset.
However, fine-tuning requires high memory and computational costs and cannot be easily conducted with devices installed in facility equipments that typically have only limited computational resources.
Therefore, a more efficient adaptation method is needed.

To address this problem, we propose AdaFlow, a Normalizing Flow-based density estimator that utilizes Adaptive Batch Normalizations (AdaBNs).
An AdaBN converts data as follows:
\begin{align}
f^{-1}_{m, \theta_{m}}(\bm{z}) =
\mbox{diag}(\bm{\gamma}) \mbox{diag}(\bm{\sigma}_{k})^{-\frac{1}{2}} (\bm{z} - \bm{\mu}_{k})) + \bm{\beta},
\end{align}
where
$\bm{\mu}_{k} \in \mathbb{R}^D$ and $\bm{\sigma}_{k} \in \mathbb{R}^D$ are vectors of mean and variance computed with the data in the $k$-th domain, respectively, and $\bm{\gamma} \in \mathbb{R}^D$ and $\bm{\beta} \in \mathbb{R}^D$ are learnable parameters shared in all domains.
The function $\mbox{diag}(\bm{\lambda})$ denotes an operator that converts $\bm{\lambda}$ into a diagonal matrix of which $(i, j)$-th entry is $(\bm{\lambda})_i$ if $i=j$, otherwise $0$.
Note that $\bm{\mu}_{k}$ and $\bm{\sigma}_{k}$ are individually calculated for each domain, whereas same $\bm{\gamma}$ and $\bm{\beta}$ are used for all $K$ domains.
By training the whole projections in this manner, $\bm{\mu}_{k}$ and $\bm{\sigma}_{k}$ alleviate the difference up to the second-order moment for each domain in the hidden layers.
In addition, adapting AdaFlow to the given $(K+1)$-th domain can be achieved by just computing AdaBNs' statistics $\bm{\mu}_{K+1}$ and $\bm{\sigma}_{K+1}$ with data sampled from that domain.

We summarize the overall procedure of pre-training and adapting AdaFlow as follows and in Fig \ref{fig:trn_flow}:
(i) pre-train AdaFlow projections with $K$ datasets by (\ref{eq:pre_train}), 
(ii) adapt the statistics of AdaBNs $\bm{\mu}'$ and $\bm{\sigma}'$ with the $(K+1)$-th dataset.

\vspace{-2.0mm}
\subsection{Examples of projection implementations }
\label{sec:implementation}

We next explain projections that can be used for implementing AdaFlow.
If each projection is easy to invert and the determinant of its Jacobian is easy to compute, exact density estimation at each data point can be easily conducted.
We introduce two projections that satisfy the above requirements.

{\bf Linear Transformation:}
Linear transformation can be used as a projection for NFs as follows:
\begin{align}
f^{-1}_{m, \theta_{m}}(\bm{z}) = \bm{W} \bm{z} + \bm{b},
\end{align}
where $\bm{W} \in \mathbb{R}^{D\times D}$ and $\bm{b} \in \mathbb{R}^{D}$ is a weight matrix and a bias vector, respectively.
The determinant of the Jacobian of this projection is $|\bm{W}|^{-1}= 1/|\bm{W}|$.
Since its computational complexity is $O(D^3)$, we reparametrize $\bm{W}$ as a LDU decomposition form
$\bm{W}=\bm{L} \mbox{diag}(\bm{d}) \bm{U}$,
where $\bm{L}$ and $\bm{U}$ is a lower and upper triangular matrix of which all diagonal elements are one, respectively, and $\bm{d} \in \mathbb{R}^{D}$.
Since $|\bm{U}|=|\bm{L}|=1$ and $|\mbox{diag}(\bm{d})|=\prod^D_i (\bm{d})_i$, the computational complexity of the determinant of the Jacobian can be reduced to $O(D)$ by using this reparametrization form.

{\bf Leaky ReLU:}
A Leaky Rectified Linear Unit (Leaky ReLU) is a module used for DNNs, defined as follows:
\begin{align}
f^{-1}_{m, \theta_{m}}(\bm{z}) = \max(\bm{z}, \alpha\bm{z}),
\end{align}
where $\alpha \in (0, 1)$ is a hyper parameter, and $\max(\bm{\lambda}^{(1)}, \bm{\lambda}^{(2)})$ is an operator that outputs element-wise maximum of $\bm{\lambda}^{(1)}$ and $\bm{\lambda}^{(2)}$, respectively.
Since Leaky ReLU is easy to invert and the determinant of its Jacobian is easy to compute, it can also be used as a projection for NFs.
The determinant of its Jacobian is $\alpha^{-\tau}$, where $\tau$ is the number of elements that are less than 0.

\vspace{-1.0mm}
\section{Experiments}
\label{sec:exp}

\vspace{-1.0mm}
\subsection{Experimental Settings}
\label{sec:exp_cond}

\subsubsection{Dataset}
\begin{figure}[ttt]
  \centering
\includegraphics[width=80mm,clip]{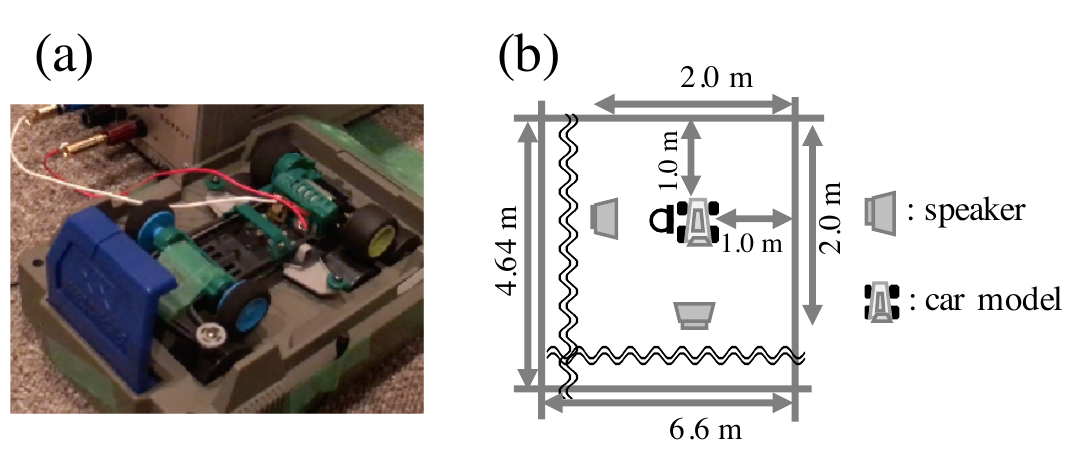}
  \caption{Photograph of toy car (left) and arrangement of toy car and loudspeakers for simulating environmental noise (right).}
  \label{fig:arrange}
  \vspace{-3.0mm}
\end{figure}

To verify the effectiveness of AdaFlow, we conducted experiments on an anomaly detection in sound (ADS) task.
For the training and test datasets, we constructed a toy-car-running sound dataset in a simulated room of a factory, as shown in Fig. \ref{fig:arrange}.
The toy cars were placed at in the room, and two loudspeakers were arranged around a toy car to emit factory noise.
For the target and noise sound, we individually collected four types of car-running sounds and four types of factory noise data emitted from two loudspeakers.
Then, $K=9$ types of pre-training datasets were generated by mixing three of the four types of car sounds and three environmental sounds at a signal-to-noise (SNR) of 0 dB.
The adaptation and test datasets were generated by mixing the remaining car sound and environmental noise at an SNR of 0 dB.
All sounds were recorded at a sampling rate of 16 kHz.

Since it is difficult to generate various types of anomalous sounds, we created synthetic anomalous sounds in the same manner as in a previous study~\cite{Koizumi_2018_IEEE_ADS}.
A part of the training dataset for the task of DCASE-2016 \cite{DCASE2016,dl_url} was used as anomalous sounds; 140 sounds including {\it slamming doors }, {\it knocking at doors }, {\it keys put on a table}, {\it keystrokes on a keyboard}, {\it drawers being opened}, {\it pages being turned}, and {\it phones ringing}) were selected.
To synthesize the test data, the anomalous sounds were mixed with normal sounds at anomaly-to-normal power ratios (ANRs) of -20 dB.
We used the area under the ROC curve (AUROC) as an evaluation metric.
We also used the negative log-likelihood (NLL).
Note that the higher AUROC, the better the model, whereas the lower NLL, the better the model.

The frame size of the discrete Fourier transformation was 512 points, and the frame was shifted every 256 samples.
The input vectors were the log amplitude spectrum of 64-dimensional Mel-filterbank outputs with a context-window size of 5.
Thus, the dimension of input vector $\bm{x}$ was $D=704$.

\vspace{-2.0mm}
\subsubsection{Comparison methods}
We compared the following models.
\begin{itemize}
  \setlength{\parskip}{0cm} 
\item AdaFlow: each model is first trained with data sampled from the nine pre-training datasets and then adapted with data sampled from the target dataset. The architecture is a sequence of linear transformation, AdaBN, leaky ReLU, linear transformation, and AdaBN. For adapting this model, the number of samples used was set to $N=10,100,1000$.
\item Normalizing Flow: each model is trained with data sampled from the nine pre-training datasets (the target dataset is not included). The architecture is a sequence of linear transformation, BN, leaky ReLU, linear transformation, and BN.
\item Normalizing Flow: a model is first trained in the same manner as above, and then fine-tuned with data sampled from the target dataset. The architecture is the same as above. For fine-tuning this model, the number of samples used was set to $N=1000$.
\item Auto-encoder: each model is trained with data sampled from the nine pre-training datasets. Since this model cannot be used for density estimation, we only evaluate AUROC. The architecture is a sequence of linear transformation (the output dimension is 128), ReLU, linear transformation (the output dimension is 64), ReLU, linear transformation (the output dimension is 128), ReLU, and linear transformation (the output dimension is 704).
\end{itemize}

\vspace{-2.0mm}
\subsection{Objective evaluations}
\label{sec:obj_eval}

The experimental results are shown in Tables 1 and 2.
From these results, we observed the following things:
\begin{itemize}
  \setlength{\parskip}{0cm} 
\item Both Normalizing Flows and AdaFlow outperformed Auto-encoder. This observation indicates the superiority of Normalizing Flows over Auto-encoder in anomaly detection.
\item AdaFlow outperformed Normalizing Flow trained with nine pre-training datasets, even when it was trained with 10 samples. This indicates the superiority of AdaFlow over non-fine-tuned Normalizing Flow.
\item The larger the amount of data used for adapting AdaFlow, the better both the metrics were. This indicates that the amount of data used for adaptation should be as large as possible.
\item AdaFlow can be adapted to a new dataset about 36 times faster than fine-tuning-based Normalizing Flow adaptation, with slight accuracy decrease. This indicates that AdaFlow is equally accurate yet much more efficient than fine-tuning-based adaptation.
\end{itemize}

\begin{table}[t]
  \begin{center}
    \caption{Results from anomaly detection experiments.}
    \vspace{-2.0mm}
    \small
    \begin{tabular}{|l||r|r|} \hline
      Method & NLL & AUROC \\ \hline \hline
      (Chance Rate) & N/A & 0.5 \\  \hline
      Norm. Flow {\small (Trained with 9 other datasets)} & 53.9 & 0.835 \\
      Auto-encoder {\small (Trained with 9 other datasets)} & N/A & 0.805 \\ \hline 
      AdaFlow {\small (Adapted with 10 samples)} & 92.4 & 0.816 \\
      AdaFlow {\small (Adapted with 100 samples)} & 21.4 & 0.875 \\
      AdaFlow {\small (Adapted with 1000 samples)} & \underline{15.3} & \underline{0.882} \\ \hline
      Norm. Flow {\small (Fine-tuned with 1000 samples)} & {\bf 13.9} & {\bf 0.887} \\ \hline
    \end{tabular}
  \end{center}
  \vspace{-7.0mm}
\end{table}

\begin{table}[t]
  \begin{center}
    \caption{Computational Time for adapting each model to the target dataset. We ran these experiments with Intel Xenon CPU (2.30GHz) on a single thread.}
    \small
    \begin{tabular}{|l||r|r|} \hline
      Method & Time [sec.]\\ \hline \hline
      Norm. Flow {\small (Fine-tuned with 1000 samples)} & 3.23 \\
      AdaFlow {\small (Adapted with 1000 samples)} & {\bf 0.09} \\ \hline
    \end{tabular}
    \vspace{-2.0mm}
  \end{center}
  \vspace{-5.0mm}
\end{table}

\vspace{-2.0mm}
\section{Application to Unpaired Cross-Domain Translation}

Though AdaFlow was originally designed for conducting density estimation on multiple domains, we demonstrate that it can be also used for the unpaired cross-domain translation problem, in which one has to train a cross-domain translation model without paired data.
We propose the unpaired cross-domain translation framework with AdaFlow in Fig. \ref{fig:trn_flow} (c).
Given a trained AdaFlow model, data belonging to one domain is first projected to the latent space with that domain's AdaBN statistics, and after that the obtained latent variable is reprojected to the data space with the target domain's AdaBN statistics.

\begin{figure}[ttt]
  \centering
  \includegraphics[width=85mm,clip]{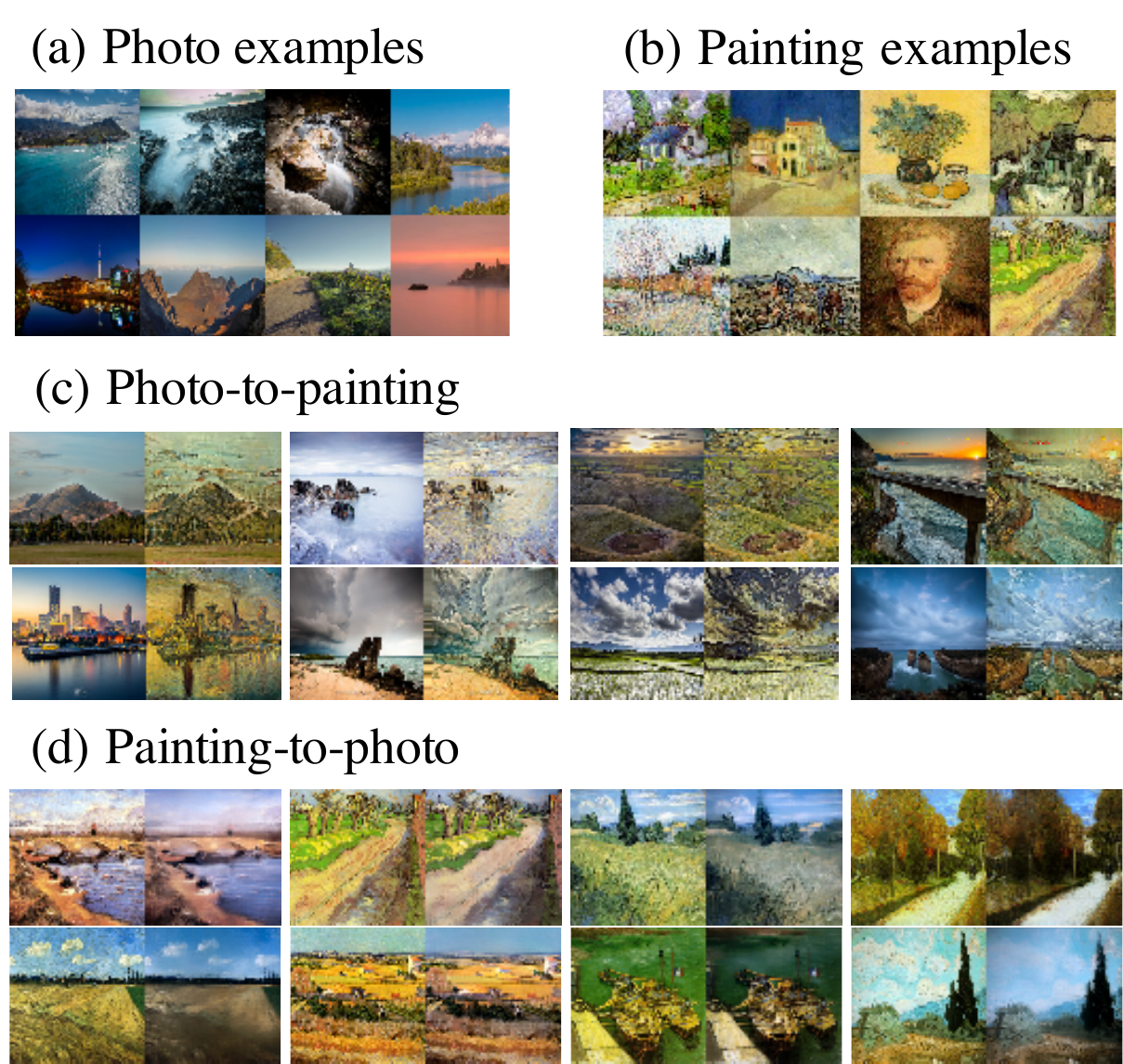}
  \vspace{-2.0mm}
\caption{Result examples of unpaired cross-domain translation.
  (a) training data examples of photos,
  (b) training data examples of paintings,
  (c) translation result examples of photo to painting, and
  (d) translation result examples of painting to photo. In both (c) and (d), input images are shown on the left side, and output images are shown on the right side.
  Best viewed in monitor.
  }
  \label{fig:crosstrans}
  \vspace{-3.0mm}
\end{figure}

We used two datasets for these experiments: the first one consisted of 400 photos, and the second one consisted of 400 paintings drawn by Van Goph.
Examples are shown in Fig. \ref{fig:crosstrans} (a) and (b).
As an architecture for AdaFlow, we employed a variant of Glow~\cite{kingma2018glow}, in which activation normalization layers are replaced with AdaBN.

The cross-domain translation results are shown in Fig. \ref{fig:crosstrans} (c, d).
We can see that unpaired cross-domain translation can be achieved via AdaFlow, even when it is trained without paired data.
These results indicate that AdaFlow can be a density-based alternative to other methods for this problem, such as CycleGAN~\cite{zhu2017unpaired}.

\vspace{-2.0mm}
\section{Conclusions}
\label{sec:cncl}

We proposed a new DNN-based density estimator called {\it AdaFlow}; a unified model of the NF and AdaBN.
Since AdaFlow can be adapted to a new domain by just adjusting the statistics used in AdaBNs, we can avoid iterative parameter update for adaptation, unlike fine-tuning. Therefore, a fast and low-computational cost domain adaptation is achieved.
We confirmed the effectiveness of the proposed method through an anomaly detection in a sound task.
We also proposed a method of applying AdaFlow to the unpaired cross-domain translation problem. We demonstrated the effectiveness of using AdaFlow for the task through cross-domain translation experiments on photo and painting datasets.

AdaFlow has the potential to resolve some problems of other important tasks.
One possible example is source enhancement~\cite{Koizumi_2018_IEEE_SE,koizumi18icassp,mogami18eusipco}.
It is known that the performance of DNN-based source enhancement is degraded when target/noise characteristics of test data are different from those of training data.
This problem is also domain-adaptation problem, thus it might be resolved by using AdaFlow.
Therefore, in the future, we plan to apply AdaFlow to other tasks including source enhancement.
\clearpage
\bibliographystyle{IEEEbib}

\end{document}